%% file: main.tex
\documentclass[letterpaper, 10 pt, conference]{ieeeconf}  
\usepackage{graphicx}
\usepackage{amsmath,amsfonts}
\usepackage{cite}
\usepackage{hyperref}
\usepackage{etoolbox}
\usepackage{graphicx} 
\usepackage{algorithmic}
\usepackage{array}
\usepackage[caption=false,font=normalsize,labelfont=sf,textfont=sf]{subfig}
\usepackage{pifont}
\usepackage{textcomp}
\usepackage[ruled,vlined]{algorithm2e}
\usepackage{stfloats}
\usepackage{multirow}
\usepackage{url}
\usepackage{verbatim}
\usepackage{graphicx}
\usepackage{color}
\usepackage{booktabs}
\usepackage[table]{xcolor}
\usepackage{threeparttable}
\usepackage{balance}
\hypersetup{hypertex=true,colorlinks=true,linkcolor=blue,anchorcolor=blue,citecolor=blue}
\usepackage{tikz}
\usepackage{caption}
\usepackage{subcaption}
\usepackage{cite}
\usepackage{balance}

\IEEEoverridecommandlockouts                         
\title{\LARGE\bf DiffAttn: Diffusion-Based Drivers' Visual Attention Prediction \\with LLM-Enhanced Semantic Reasoning}

\author{Weimin Liu$^{1}$, Qingkun Li$^{2*}$, Jiyuan Qiu$^{3}$, Wenjun Wang$^{1*}$, Joshua H. Meng$^{4}$
\thanks{* Corresponding authors: Wenjun Wang and Qingkun Li.}
\thanks{$^{1}$Weimin Liu is with School of vehicle and Mobility, Tsinghua University, Beijing, China {\tt\small lwm23@mails.tsinghua.edu.cn}}
\thanks{$^{2}$Qingkun Li is with Beijing Key Laboratory of Human-Computer Interaction, Institute of Software, Chinese Academy of Sciences, Beijing, China {\tt\small qingkun.li.thu@gmail.com}}
\thanks{$^{3}$Jiyuan Qiu is with Remote Sensing and Earth Observation Laboratory, University of Copenhagen, Copenhagen K, Denmark {\tt\small jiqi@ign.ku.dk}}
\thanks{$^{1}$Wenjun Wang is with School of vehicle and Mobility, Tsinghua University, Beijing, China {\tt\small wangxiaowenjun@tsinghua.edu.cn}}
\thanks{$^{4}$Joshua H. Meng is with California PATH, University of California, Berkeley, CA, USA {\tt\small hdmeng@berkeley.edu}}
}

\begin{document}
\maketitle
\thispagestyle{empty}
\pagestyle{empty}

\begin{abstract}
\par Drivers' visual attention provides critical cues for anticipating latent hazards and directly shapes decision-making and control maneuvers, where its absence can compromise traffic safety. To emulate drivers' perception patterns and advance visual attention prediction for intelligent vehicles, we propose DiffAttn, a diffusion-based framework that formulates this task as a conditional diffusion-denoising process, enabling more accurate modeling of drivers' attention. To capture both local and global scene features, we adopt Swin Transformer as encoder and design a decoder that combines a Feature Fusion Pyramid for cross-layer interaction with dense, multi-scale conditional diffusion to jointly enhance denoising learning and model fine-grained local and global scene contexts. Additionally, a large language model (LLM) layer is incorporated to enhance top-down semantic reasoning and improve sensitivity to safety-critical cues. Extensive experiments on four public datasets demonstrate that DiffAttn achieves state-of-the-art (SoTA) performance, surpassing most video-based, top-down-feature-driven, and LLM-enhanced baselines. Our framework further supports interpretable driver-centric scene understanding and has the potential to improve in-cabin human-machine interaction, risk perception, and drivers' state measurement in intelligent vehicles.
\end{abstract}

\input{sections/intro}
\input{sections/method}

\input{sections/exp}

\section{Conclusion}
\label{sec::conclusion}
\par We propose DiffAttn, a diffusion-based framework for drivers' visual attention prediction that formulates the task as a conditional diffusion-denoising process aligned with Gaussian-like human attention patterns. A Swin Transformer encoder with a multi-scale conditional decoding design captures both local details and global scene context, while an LLM layer enhances top-down semantic representations for safety-critical cues. Experiments on four public datasets demonstrate SoTA performance of our method, surpassing multiple video-based, top-down-feature-driven, and LLM-enhanced methods. Our findings offer prospects for modeling drivers' visual behavior and provide insights into the application possibilities of visual attention prediction for intelligent human-machine interaction and drivers' state measurement.

\section{Acknowledgment}
\par This work was jointly supported by National Natural Science Foundation of China under Grant 52502518, CAS Major Project under Grant RCJJ-145-24-14, the Open Fund Project of State Key Laboratory of Intelligent Green Vehicle and Mobility under Grant KFY260407 and Tsinghua University-Toyota Joint Research Center for AI-Technology of Automated Vehicle under Grant TTAD-2025-05.

\bibliographystyle{IEEEtran}
\bibliography{ref}
\end{document}

%% file: sections/intro.tex
\section{Introduction}
\label{sec:intro}
\par As autonomous driving systems and intelligent vehicular algorithms have advanced significantly in recent years, drivers now have more opportunities to engage in non-driving-related tasks (NDRTs) during prolonged and monotonous autonomous driving, where their gazes are no longer required to be continuously fixed on the road \cite{li2022understanding}. Under such conditions, accurate measurement and assessment of drivers’ visual attention distribution becomes critically important for ensuring the safety and reliability of autonomous vehicles, especially in scenarios requiring human-machine cooperation or take-over \cite{liu2024deep}. Reliable attention measurement not only provides quantitative indicators of drivers’ cognitive states, but also serves as a fundamental component for driver monitoring systems, risk evaluation, and adaptive human-vehicle interfaces in automated driving \cite{liu2023literature}.
\par Recent advancements in vision-based models for intelligent vehicles, such as object detection \cite{qiu2024etformer} and semantic segmentation \cite{qiu2024evsmap}, have shown progress in individual tasks, yet they struggle to identify crucial visual cues and understand scene risks involved in traffic environment like experienced drivers do in case of an emergency \cite{baee2021medirl}. In comparison to machine intelligence, humans are capable of quickly detecting the most relevant stimuli, and locating potential hazards in complex situations through visual attention \cite{deeptake}. In situations where dynamic driving task (DDT) execution relies heavily on vision for scene perception and understanding, drivers' visual attention is essential for perceiving the traffic environment and interacting with traffic participants, since visual attention provides crucial cues for their intended control maneuvers and accident avoidance capabilities. Overlooking latent hazards can pose threats to traffic safety and potentially result in accidents and casualties \cite{li2023latent}. For traffic safety, drivers' visual attention prediction by mimicking human drivers' visual behavior and attention mechanisms, could be greatly beneficial in supporting autonomous driving, assessing drivers' states, and delivering hazard warnings.
\begin{figure}[t]
    \centering
    \includegraphics[width=1\columnwidth]{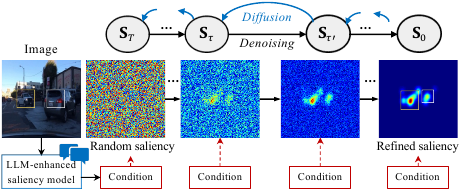}
    \caption{Overview of the proposed LLM-enhanced, conditional diffusion-based drivers' visual attention modeling method \textbf{DiffAttn}.}
    \label{fig::overview}
\end{figure}
\par Drivers' visual attention can be categorized into bottom-up and top-down mechanisms. Bottom-up control is data-driven and guided by salient objects or areas in the driving scene that stand out against the background due to image-based conspicuities. These elements attract or even distract drivers' attention, such as billboards, advertisements, or vehicles driving in the opposite lane, which are less safety-critical \cite{itti2000saliency}. Top-down attention, however, is task-driven and goal-oriented where factors including experience, knowledge, memory, and expectations could prompt and guide drivers to focus on objects or events relevant to the driving-task-related information and allocate less attentional resources to stimuli that are irrelevant. For instance, drivers mostly fixate their attention on vanishing points to get a broader view of the road ahead \cite{review}. In complex dynamic driving environments, both bottom-up and top-down factors continuously evolve and compete for drivers' visual attentional resources. Therefore, both types of factors should be considered when modeling drivers' visual attention when executing DDT.
\par In this work, we aim to propose a diffusion-based framework to model human-like visual attention pattern without additionally depending on top-down features like temporal information, segmentation maps or optical flows. The main contributions of this study are summarized as follows.
\par (1) We propose \textbf{DiffAttn}, a diffusion-based framework that formulates drivers' visual attention prediction as a conditional denoising process, which naturally aligns with the Gaussian-like spatial distribution of human gazes, providing both theoretical coherence and superior performance.
\par (2) To capture local and global contexts in traffic scenes, we adopt Swin Transformer as encoder and a decoder that combines channel-attentive feature fusion with dense multi-scale conditioning and prediction, enabling effective denoising with fine-grained details and holistic scene awareness.
\par (3) To enhance top-down feature representations without explicitly introducing top-down modalities, we incorporate a LLM layer into the saliency decoder of the network, enabling the model to better reason about semantic context across scales and strengthen its sensitivity to safety-critical cues.
\par (4) Extensive experiments on four public datasets demonstrate SoTA performance of our method, surpassing video-based approaches, top-down-feature-driven methods, and even some LLM-enhanced baselines. Beyond benchmark results, our framework also promotes interpretable driver-centric scene understanding and has potential to support safer human-machine interaction in intelligent driving systems.

\subsection{Related Works}
\par\textbf{CNN-based approaches.} In the studies of Ji $\textit{et al.}$ \cite{DPSNN} and Deng $\textit{et al.}$ \cite{CDNN}, the network models were designed in a U-Net pattern, where cascaded CNN layers in encoder extract image features and the decoder gradually upsamples feature maps and make final predictions. To bring more temporal dependencies and top-down features into the network, Xia $\textit{et al.}$ \cite{HWS} adopted ConvLSTM as temporal processing module to predict drivers' attention in critical situations from video clips. By doing this, information of previous fixation locations could flow along sequence which benefits sequential attention inferences. Fang $\textit{et al.}$ \cite{DADA} used segmentation labels as top-down cues to facilitate task-specific attention allocation in their network SCAFNet to reason semantic-induced scene variation in drivers' visual attention predictions. However, the improvements in model performance were built on sacrifice of lightweight architecture and computation expenses, as network composes more complicated modules for bridging relationships among cues. 
\par\textbf{Transformer-based approaches.} Transformer was first utilized in natural language processing studies. To leverage Transformer's power in long-range dependency modeling, Vision-Transformer (ViT) \cite{dosovitskiy2020image} divides an RGB image into patches, flattens them, and treats the image as sequential data for Transformer input. Zhao $\textit{et al.}$ \cite{gate} proposed Gate-DAP and explored the network connection gating mechanism for driver attention prediction to boost prediction performance through learning the importance of input top-down features like segmentation maps and optical flows. Although Gate-DAP shows accurate prediction performance on DADA-2000 and BDD-A datasets, the ViT-based backbone design might hinder network performance in modeling drivers' foveal vision which constitutes local information of visual attention.
\par Most existing studies on drivers’ visual attention prediction directly map RGB images $\mathbf{I}$ to attention maps $\mathbf{S}$ via supervised training. Although effective, these approaches treat attention prediction as a deterministic regression task that estimates the conditional expectation $\mathbb{E}(\mathbf{S}|\mathbf{I})$, overlooking the probabilistic and spatially distributed nature of human gaze. Diffusion-based methods instead frame attention prediction as a conditional generative process, gradually refining a noisy map into a realistic attention distribution $p(\mathbf{S}|\mathbf{I})$.

%% file: sections/method.tex
\section{Method}
\label{sec::framework}
\subsection{Task Definition and Motivation}
\par Drivers' visual attention prediction seeks to forecast the intensity and distribution of visual attention on salient areas within the driving scene. This is achieved by predicting a saliency map $\hat{\mathbf{S}}\in\mathbb{R}^{1\times H\times W}$, which indicates the probability of fixation occurrence. This formulation is subtracted to the observation that when humans allocate visual attention to a scene, foveal vision provides the highest resolution of fine-grained local visual information and allows for acuity and contrast sensitivity around fixations \cite{stewart2020review}. Meanwhile, peripheral vision provides non-detailed, coarse-grained, texture-like yet long-range visual information \cite{stewart2020review}. Consequently, a saliency map is normally adopted to depict the perceptual spatial interactions of both local and non-local information, as well as the intertwined foveal and peripheral vision. 
\par To mimic drivers' visual attention allocation pattern, the groundtruth saliency map $\mathbf{S}$ is generated by conducting Gaussian filtering on a binary fixation map $\mathbf{F}$, where $\mathbf{F}(x_0,y_0)=1$ when $(x_0,y_0)$ has valid fixation from the driver. The formulation of groundtruth saliency map is given as,
\begin{equation}
    \mathbf{S}(x,y)=\sum_{i=1}^W\sum_{j=1}^H\mathbf{F}(i,j)\cdot\mathbf{G}(x-i,y-j),
\end{equation}
where $\mathbf{G}(x,y)=\frac{1}{2\pi\sigma_x \sigma_y}\exp\left[-\frac{1}{2}\left(\frac{x^2}{\sigma_x^2} +\frac{y^2}{\sigma_y^2}\right)\right]$, $\sigma_x$ and $\sigma_y$ implies standard deviations of Gaussian kernel $\mathbf{G}$ in $x$- and $y$-axis to represent parafoveal and the peripheral area around fixations, respectively. Typically, $\sigma_x=\sigma_y$. The final groundtruth saliency map is generated with min-max normalization, indicating the probability of visual attention falling within the vicinity of fixations. 
\par
\begin{figure*}[t]
    \centering
    \includegraphics[width=2\columnwidth]{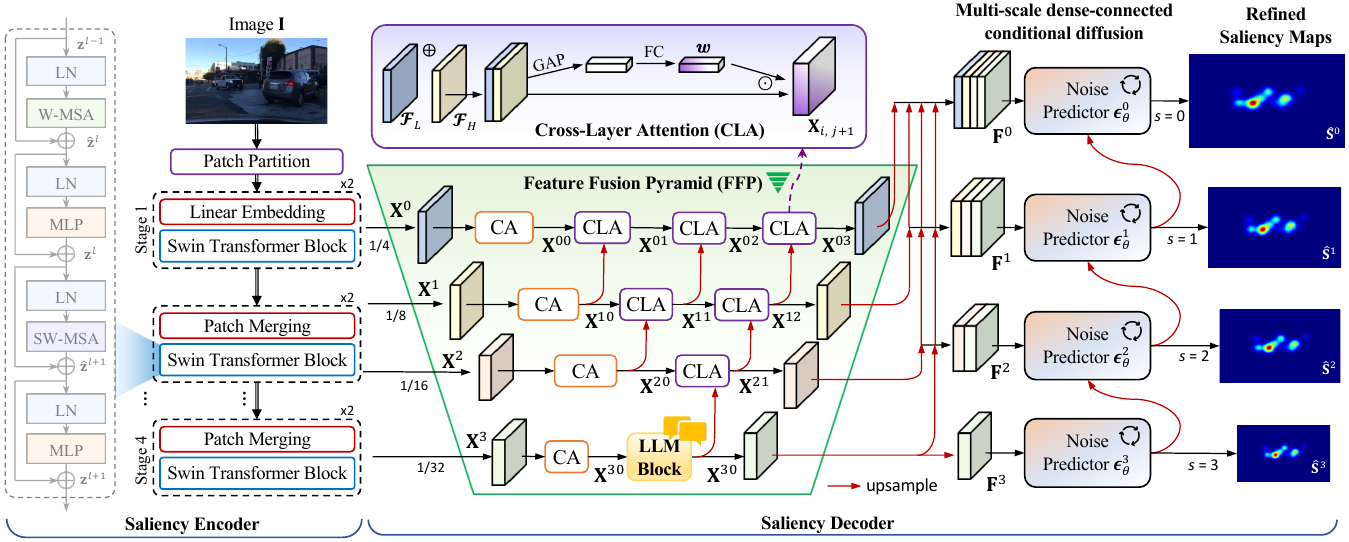}
    \caption{\textbf{DiffAttn} architecture overview. For saliency encoder, we adopt SwinT-Base \cite{swinT} pretrained on ImageNet. The decoder is designed with an LLM-enhanced feature fusion pyramid (FFP), which bridges the encoder outputs, and a multi-scale dense-connected conditional diffusion module, where feature maps produced by FFP are densely connected and serve as conditioning signals for noise learning in the diffusion process. The noise predictors generate saliency maps at multiple scales, which are all supervised with groundtruth saliency maps. Saliency map generated at $s=0$ during testing.}
    \label{fig::network}
\end{figure*}

 In this work, we proposed DiffAttn, where the network takes color image $\mathbf{I}\in\mathbb{R}^{3\times H\times W}$ as inputs only, and outputs saliency maps $\hat{\mathbf{S}}^s\in\mathbb{R}^{1\times\frac{H}{2^s}\times \frac{W}{2^s}}$ at four scales $\mathcal{S}=\left\{s\vert0,1,2,3\right\}$ to constitute training losses with groundtruth. By model inference, only saliency map at input resolution $\hat{\mathbf{S}}^0$ would be used for evaluation. Overview of the framework of our proposed network is shown in Fig. \ref{fig::network}. 

\subsection{Preliminaries of Diffusion Model}
\par Diffusion models have demonstrated remarkable generative capabilities across a variety of tasks. The Denoising Diffusion Probabilistic Model (DDPM) \cite{ddpm} introduced a framework that leverages Markovian processes in both forward and reverse stages. The Denoising Diffusion Implicit Model (DDIM) \cite{ddim} improved DDPM by utilizing non-Markovian chains for faster sampling speed. Typically, diffusion models are categorized into unconditional and conditional types. Unconditional models aim to directly approximate the data distribution, whereas conditional models focus on generating data under specific conditions.
\par In conditional diffusion models, target data distribution $\mathbf{x}_0$ is transformed into a noisy sample $\mathbf{x}_T$ through a sequence of conditional probabilities $q(\mathbf{x}_\tau|\mathbf{x}_0)$ as, 
\begin{equation}
    q(\mathbf{x}_\tau|\mathbf{x}_0):=\mathcal{N}(\mathbf{x}_\tau;\sqrt{\overline{\alpha}_\tau}\mathbf{x}_0,(1-\overline{\alpha}_\tau)\mathbf{I}),
\end{equation}
\begin{equation}
    \overline{\alpha}_\tau:=\prod_{t=0}^\tau\alpha_t=\prod_{t=0}^\tau(1-\beta_t),
\end{equation}
\begin{equation}
    \mathbf{x}_\tau=\sqrt{\overline{\alpha}_\tau}\mathbf{x}_0+\sqrt{1-\overline{\alpha}_\tau}\boldsymbol{\epsilon},~\boldsymbol{\epsilon}\sim\mathcal{N}(\boldsymbol{0},\mathbf{I}),
\end{equation}
where $q$ represents the forward noise introduction process. $\beta_t$ denotes the noise variance schedule as defined in DDPM \cite{ddpm}, and $\mathcal{N}$ implies Gaussian distribution. 
\par In the denoising process, a noise predictor $\boldsymbol{\epsilon}_\theta$ learns to reverse the diffusion process and iteratively recover $\mathbf{x}_0$ under the guidance of visual conditions $\mathbf{c}$ as,
\begin{equation}
    p_\theta(\mathbf{x}_{\tau-1}|\mathbf{x}_\tau,\mathbf{c}):=\mathcal{N}(\mathbf{x}_{\tau-1};\boldsymbol{\mu}_\theta(\mathbf{x}_\tau,\tau,\mathbf{c}),\sigma_\tau^2\mathbf{I}),
\end{equation}
\begin{equation}
    \boldsymbol{\mu}_\theta(\mathbf{x}_\tau,\tau,\mathbf{c})=\frac{1}{\sqrt{\alpha_\tau}}(\mathbf{x}_\tau-\frac{\beta_\tau}{\sqrt{1-\overline{\alpha}_\tau}}\boldsymbol{\epsilon}_\theta(\mathbf{x}_\tau,\tau,\mathbf{c})),
\end{equation}
where $\sigma_\tau^2$ indicates transition variance.

\subsection{Saliency Encoder}
\par Encoder plays an essential role in feature representation from model inputs. While CNNs have been widely used in prior works such as \cite{CDNN}, their local receptive fields limit long-range contextual modeling. Given that drivers’ attention involves both local visual cues and global contextual information, in this work, we adopt Swin Transformer \cite{swinT} as backbone, as its hierarchical shifted window attention design allows it to effectively model both local details and global context. The configuration of Swin Transformer block is shown in Fig. \ref{fig::network}. Hierarchical outputs of Swin Transformer-based saliency encoder are $\mathbf{X}^0\in\mathbb{R}^{C_e\times\frac{H}{4}\times\frac{W}{4}}$, $\mathbf{X}^1\in\mathbb{R}^{2C_e\times\frac{H}{8}\times\frac{W}{8}}$, $\mathbf{X}^2\in\mathbb{R}^{4C_e\times\frac{H}{16}\times\frac{W}{16}}$, and $\mathbf{X}^3\in\mathbb{R}^{8C_e\times\frac{H}{32}\times\frac{W}{32}}$.

\subsection{Saliency Decoder}
\par DiffAttn decoder takes multi-scale features from the encoder and hierarchically upsamples these skip-connected feature maps from deep to shallow layers to form saliency predictions. As shown in Fig. \ref{fig::network}, DiffAttn saliency decoder consists of Feature Fusion Pyramid (FFP) module with LLM-based semantic enhancement along skip connection path, and multi-scale dense-connected conditional diffusion. 
\par\textbf{Feature fusion pyramid.} In this work, we integrate a feature fusion pyramid (FFP) between the encoder and decoder to enhance multi-scale feature aggregation, preserve spatial details, aid the recovery of information lost during downsampling, and strengthen the complementary representation of features across levels. As illustrated in Fig. \ref{fig::network}, FFP employs a channel attention (CA) mechanism to refine feature representations and facilitate cross-layer interactions.

\par Output from the saliency encoder at each scale $\{\mathbf{X}^{ii}\}_{i=0}^4$ is first processed by a channel attention module, followed by a series of cross-layer attention (CLA) modules. Design of CLA is to enhance encoder features and facilitate hierarchical information flow within FFP. CLA is formulated by channel attention on concatenated lower- and higher-level features,
\begin{equation}
    \mathbf{X}^{i,j+1}=\text{CA}\left([\mathbf{X}^{i,j}, u(\text{ELU}(\kappa(\mathbf{X}^{i+1,j})))]\right),
\end{equation}
where $\mathbf{X}^{i,j+1}$ denotes the intermediate output of each CLA cell. $\mathbf{X}^{i,j}$ represents the lower-level feature. The higher-level feature $\mathbf{X}^{i+1,j}$ is first processed by a convolution operation $\kappa$ with ELU activation, and then upsampled by $u$. FFP produces fused feature maps $\mathbf{X}^{22}$, $\mathbf{X}^{13}$, and $\mathbf{X}^{03}$ at resolutions $(\tfrac{H}{4}, \tfrac{W}{4})$, $(\tfrac{H}{8}, \tfrac{W}{8})$, and $(\tfrac{H}{16}, \tfrac{W}{16})$, respectively. 
\par\textbf{LLM-based semantic enhancement.} LLMs provide strong capabilities for capturing high-level semantics, reasoning about contextual relationships, and transferring knowledge from large-scale training. For drivers’ visual attention prediction, these capabilities are particularly valuable because gaze behavior is influenced not only by bottom-up visual saliency but also by top-down semantic understanding. Recent work such as SalM$^2$ \cite{salm2} leverages the CLIP \cite{clip} model to enrich semantic representations of driving scenes. In this work, we adopt the strategy proposed in LLM4Seg \cite{llm4seg} to further leverage the capability of LLMs in enhancing visual semantic understanding, thereby improving the modeling of drivers’ visual behavior, particularly top-down attention, which remains a challenge that most existing studies have not explicitly addressed. To this end, we introduce an additional LLM layer $\mathcal{F}_\text{LLM}$ along the skip-connection path from the saliency encoder to the decoder at the deepest level. Concretely, the feature map $\mathbf{X}^{30}$ is first reshaped and flattened into $(8C_e,\tfrac{H \times W}{32^2})$, linearly projected by $\phi$, passed through the LLM layer $\mathcal{F}_\text{LLM}$, subsequently linearly projected by $\psi$, and finally reshaped back to $(8C_e,\tfrac{H}{32}, \tfrac{W}{32})$. The two trainable linear projectors $\phi$ and $\psi$ bridge the discrepancy between visual patch tokens and LLM's text embedding space. This process can be expressed as,
\begin{equation}
\begin{aligned}
    &~~~\mathbf{y}^{30}=\phi\left(\text{Reshape~\&~Flatten}\left(\mathbf{X}^{30}\right)\right),\\
    &~~~~\mathbf{X}^{30}\leftarrow\text{Reshape}\left(\psi\left(\mathcal{F}_\text{LLM}\left(\mathbf{y}^{30}\right)\right)\right).\\
\end{aligned}
\end{equation}
\begin{figure}[h]
    \centering
    \includegraphics[width=1\columnwidth]{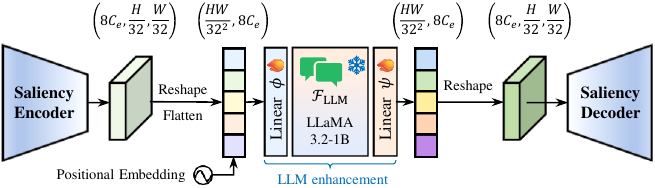}
    \caption{Network architecture of LLM-based semantic enhancement.}
    \label{fig::LLM}
\end{figure}
\par In this work, LLM enhancement is applied only at the deepest scale level, both to reduce computational cost and to ensure that its benefits can be propagated to higher levels through feature fusion and dense cross-scale connections, as discussed in the following subsection.
\par\textbf{Multi-scale dense-connected conditional groundtruth diffusion.} In this work, we adopt groundtruth driver attention map $\mathbf{S}_g$ as target data distribution. The generation of a noisy sample $\mathbf{S}_\tau$ can be written as,
\begin{equation}
    q(\mathbf{S}_\tau|\mathbf{S}_g):=\mathcal{N}(\mathbf{S}_\tau|\sqrt{\overline{\alpha}_\tau}\mathbf{S}_g,(1-\overline{\alpha}_\tau)\mathbf{I}).
\end{equation}
\par
\begin{figure}[h]
    \centering
    \includegraphics[width=0.85\columnwidth]{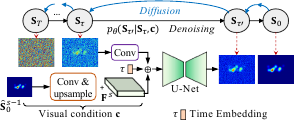}
    \caption{Network architecture of multi-scale conditional diffusion.}
    \label{fig::diffusion}
\end{figure}

 The subsequent noise modeling process is realized through a U-Net network. As shown in Fig. \ref{fig::diffusion}, the network takes as input the summation of the visual condition $\mathbf{c}$, the current time embedding, and the noisy sample at time step $\tau$, and outputs the denoised sample corresponding to time step $\tau'$. Multi-scale prediction strategy, which has been proven effective in many vision tasks such as semantic segmentation and depth estimation, where each scale has an individual noise predictor, denoted as $\boldsymbol{\epsilon}_\theta^s$. Leveraging predictions at multiple resolutions allows the model to account for scale variations, where finer scales emphasize small, localized cues, while coarser scales highlight the centers of larger salient regions. The visual condition on each output scale $s$ is calculated by hierarchically upsampling and concatenating feature outputs of FFP from lower level for feature representation enhancement, as well as adding refined saliency output from lower level. An example for the calculation of visual condition at bottom $(s=3)$ and top level $(s=0)$ are as,
\begin{equation}
    \mathbf{F}^3=\text{upsample}_{\times 2}(\mathbf{X}^{30}),~ \mathbf{c}^3=\text{ELU}(f^3(\mathbf{F}^3)),
\end{equation}
\begin{equation}
    \mathbf{F}^0=\left[
            u_{\times 2}(\mathbf{X}^{03});
            u_{\times 3}(\mathbf{X}^{12});
            u_{\times 4}(\mathbf{X}^{21});    
            u_{\times 5}(\mathbf{X}^{30})\right],
\end{equation}
\begin{equation}
    \mathbf{c}^0= \text{ELU}\left(f^0\left(\mathbf{F}^0\right)\right)+u\left(g^0\left(\hat{\mathbf{S}}^{s-1}\right)\right),
\end{equation}
where $\mathbf{c}^s$ and $(f^s,g^s)$ denote the visual condition and 2D convolution operations at scale $s$, respectively.
\par The final driver attention maps are generated by applying a sigmoid activation to the denoised outputs at each scale.

\subsection{Loss Function}
\label{sec::loss}
\par The overall loss function, averaged over all scales, is composed of binary cross entropy (BCE) loss, KL-Divergence (KLD) loss and diffusion denoising (DD) loss,
\begin{equation}
    \mathcal{L}=\frac{1}{|\mathcal{S}|}\sum_{s\in\mathcal{S}}\lambda_1\mathcal{L}_\text{BCE}^s+\lambda_2\mathcal{L}_\text{KLD}^s+\lambda_3\mathcal{L}_\text{DD}^s,
\end{equation}
where $\lambda$ implies loss weights. The direct output of the saliency map at each scale is upsampled to input resolution for the calculation of each respective loss component.
\par In line with DDIM \cite{ddim}, diffusion denoising loss supervises the latent saliency at sampled time step after refinement by reversing the diffusion process. It could be written as,
\begin{equation}
    \mathcal{L}_\text{DD}^s=\Vert\boldsymbol{\epsilon}^s-\boldsymbol{\epsilon}_\theta^s(\mathbf{S}_\tau^s,\tau,\mathbf{c}^s)\Vert^2.
\end{equation}

%% file: sections/exp.tex
\section{Experiments}
\label{sec::experiments}
\subsection{Datasets}
\par\textbf{TrafficGaze} \cite{CDNN} contains 16 video clips with resolution $(1280,720)$ and recorded on urban roads in China. Each image has 28 gaze providers for fixation collections in the lab. 10, 2 and 4 videos are used for train, validation and test. 
\par\textbf{DADA-2000} \cite{DADA}. Driver Attention and Accident Dataset contains 2000 recorded accident videos on crowded city roads from various locations worldwide, sourced from websites. The video clips have a resolution of $(1584, 660)$, and each frame is annotated by five gaze providers.
\par\textbf{BDD-A} \cite{BDDA}. Berkeley DeepDrive Attention dataset contains 926, 203 and 306 video clips for train, validation and test of visual attention prediction. Images are with resolution $(1280,720)$, where each frame has 4 providers for gaze collection in the lab. The driving videos include braking events and were recorded in busy urban areas in the US.
\par\textbf{DrFixD-rainy} \cite{drfixd-rainy} contains 16 traffic driving videos in rainy conditions, where image frames are with resolution $(1280,720)$. Each frame has 30 in-lab gaze providers. 10, 2
and 4 videos are assigned for training, validation and test.

\subsection{Implementations Details}
\par We implement the proposed method \textbf{DiffAttn} in PyTorch and train it on a single NVIDIA RTX 3090 GPU. The model is optimized using AdamW with a batch size of 18, a learning rate of $1\times10^{-5}$, and a weight decay of $0.001$. All input RGB images are resized to $(192,320)$, with random color jittering and horizontal flipping applied for data augmentation. The loss function employs weighting coefficients of $\lambda_1=1$ and $\lambda_3=0.001$ for all datasets, while $\lambda_2$ is set to $\{1,0.2,0.1,1\}$ for TrafficGaze, DADA-2000, BDD-A, and DrFixD-rainy, respectively. Update rule of DDIM \cite{ddim} was selected for the sampling. Diffusion step $T_i$ is set to 300 on all scales for all datasets. Denoising steps $T_e$ are set to $\{12,16,15,12\}$ on all scales for TrafficGaze, DADA-2000, BDD-A, and DrFixD-rainy, respectively. For LLM enhancement, we employ the frozen 15$^\text{th}$ layer of LLaMA3.2-1B \cite{llama} as LLM layer $\mathcal{F}_\text{LLM}$. The linear projections within LLM enhancement module are set to $(C_e,2028)$ and $(2048,C_e)$, respectively.
\par The evaluation employs KLD, SIM, CC, NSS and AUC-J metrics to assess drivers' visual attention prediction.

\subsection{Experiment Results}
\input{tables/result_all}
\begin{figure*}[t]
    \centering
    \includegraphics[width=2\columnwidth]{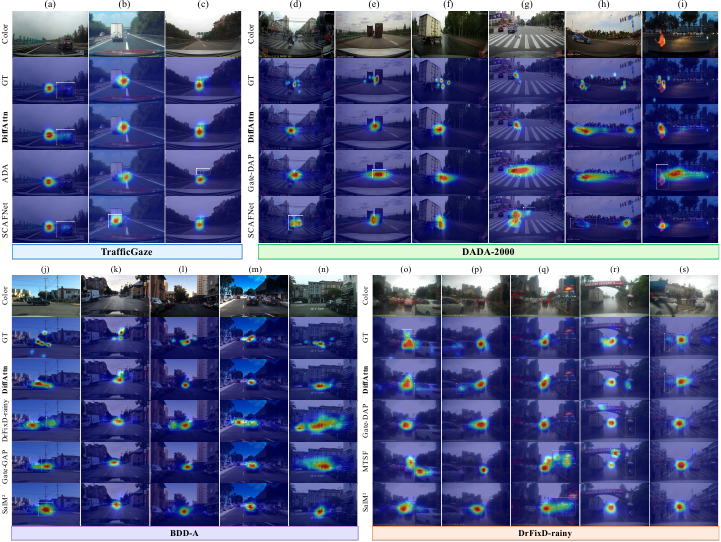}
    \caption{\textbf{Qualitative results on TrafficGaze:} (a) Surrounding vehicle driving in right lane; (b) Changing to right lane with a truck ahead; (c) Straight driving with a traffic sign ahead. \textbf{Qualitative results on DADA-2000:} (d) Motorcycle crossing; (e) Two trucks ahead; (f) Nearby truck changing lane ahead; (g) Pedestrian crossing; (h) Turning right with collision risk involving a taxi; (i) Pedestrian running in front of ego-vehicle. \textbf{Qualitative results on BDD-A:} (j) Vehicle crossing ahead; (k) Straight driving with a traffic light ahead; (l) Driving past parked cars; (m) Lane change with a braking vehicle ahead; (n) Approaching a STOP line. \textbf{Qualitative results on DrFixD-rainy:} (o) Entering a main road with congested traffic; (p) Nearby left vehicle changing into the ego lane; (q) Driving through a green light; (r) Driving on a rural road with a bicyclist on the right; (s) Pedestrians standing at the roadside.}
    \label{fig::result}
\end{figure*}

\par The proposed \textbf{DiffAttn} outperforms 15 baseline methods across all four benchmarks, as summarized in Table \ref{tab::result_all} and visualized in Fig. \ref{fig::result}. Qualitative comparisons in Fig. \ref{fig::result} demonstrate DiffAttn's superiority in capturing both bottom-up salient regions and top-down semantic features. In normal driving scenarios, DiffAttn exhibits enhanced top-down reasoning capabilities. For instance, in Fig. \ref{fig::result}(a)(b), while baseline models like ADA restrict attention to the forward view, DiffAttn additionally attends to safety-critical elements such as overtaking vehicles and traffic signs. In complex and hazardous situations, DiffAttn shows precision in localizing latent risks. For instance, in accident-prone scenarios shown in Fig. \ref{fig::result}(d)(g)(h), DiffAttn reliably detects high-risk agents like crossing bicyclists and pedestrians, while simultaneously maintaining awareness of the intended path and contextual cues. By contrast, SCAFNet and Gate-DAP either miss critical hazards or produce attention artifacts. At intersections and under adverse conditions, DiffAttn closely aligns with human drivers' attention patterns. As shown in Fig. \ref{fig::result}(k)(n)(q), DiffAttn jointly focuses on traffic lights, STOP signs, and roadway geometry, whereas competing models often neglect these top-down semantic features or scatter attention to irrelevant background elements. These results highlight DiffAttn's ability to balance immediate local hazards with higher-level driving semantics, producing attention maps with richer spatial details and stronger consistency with human gaze patterns.

\subsection{Ablation Studies}
\par This subsection presents the results of ablation studies on our experimental setting, where most experiments were conducted on two challenging datasets, DADA-2000 and BDD-A, to validate both the effectiveness and the underlying rationale of our proposed algorithm.

\par\textbf{Ablation study on LLM-based semantic enhancement.} Table \ref{tab::ablation_LLM} reports results on the BDD-A and DADA-2000 datasets when the LLM layer is removed or replaced with the 28$^\text{th}$ layer of DeepSeek-R1-Distill-Qwen-1.5B \cite{guo2025deepseek}, following LLM4Seg \cite{llm4seg}. 
Removing the LLM layer consistently degrades performance, indicating that it provides essential semantic cues for accurate attention prediction while introducing only a small parameter overhead. In addition, loading pretrained weights from LLaMA or DeepSeek consistently improves performance, and using a frozen layer outperforms a trainable one. This suggests that the pretrained LLM’s ability to model long-range dependencies and capture rich global semantic representations that cannot be reliably fine-tuned from the limited, task-specific saliency data. Furthermore, despite from text to visual modality, LLaMA achieves slightly better results than DeepSeek variant, likely due to the rich, abstract semantic representations it encodes from large-scale text corpora, which provide high-level guidance for attention allocation in visual scenes.
\input{tables/ablation_LLM}
\par
\begin{figure}[t]
    \centering
    \includegraphics[width=1\columnwidth]{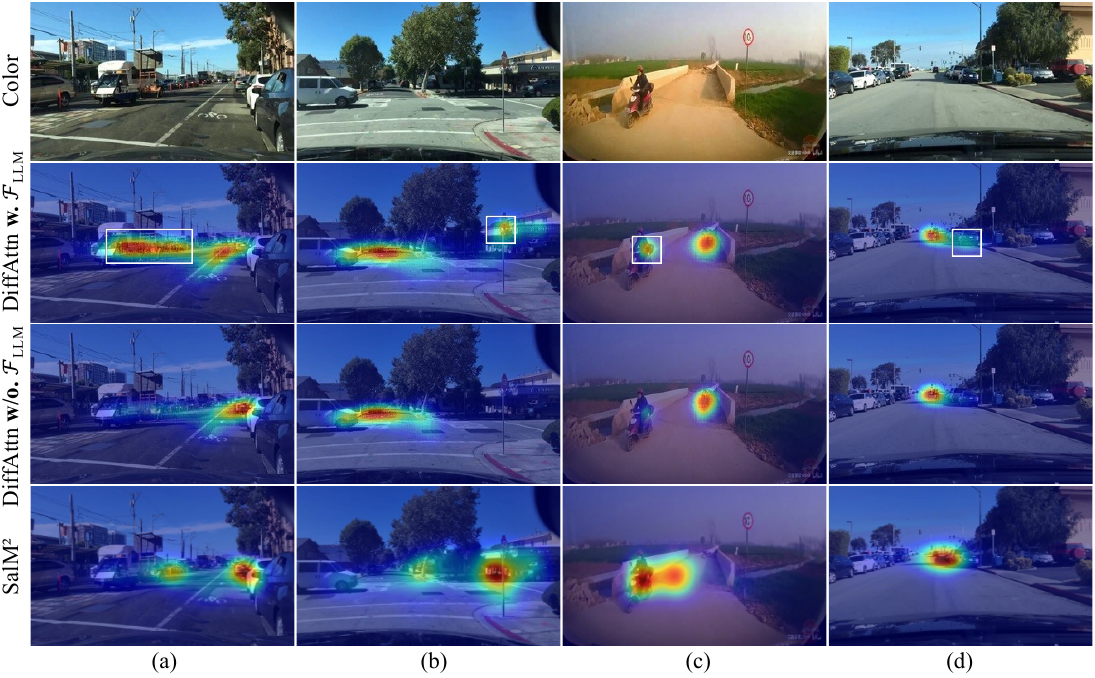}
    \caption{Qualitative comparison of saliency prediction results: with LLM enhancement (second row), without LLM enhancement (third row), and CLIP-based method SalM$^2$ \cite{salm2} (last row).}
    \label{fig::LLM_compare}
\end{figure}

 In Fig. \ref{fig::LLM_compare}, we present a qualitative comparison of saliency prediction results with different LLM options. Specifically, in Fig. \ref{fig::LLM_compare}(a), when the ego-vehicle is steering left and entering the main road, the proposed LLM-enhanced variant allocates attention not only to the distant point along the main road, which is primarily focused on by the non-enhanced variant or the CLIP-enhanced SalM$^2$ method, but also to the vehicles parked on the opposite roadside. Meanwhile, in Fig. \ref{fig::LLM_compare}(b), the LLM-enhanced variant attends strongly to the driving trajectory of the vehicle crossing ahead, while also noticing a road sign at the intersection to avoid potential conflicts or violations of traffic rules. These examples indicate that our proposed LLM-enhanced method could better comprehend traffic scenes and allocate more top-down attention to event-driven features that are critical for driving safety. In addition, as reported in \cite{salm2}, SalM$^2$ model requires square-shaped inputs due to constraints of CLIP model, whereas our proposed LLM-enhancement method is considerably more flexible with respect to input resolution.
\par
\begin{figure}[h]
    \centering
    \includegraphics[width=1\columnwidth]{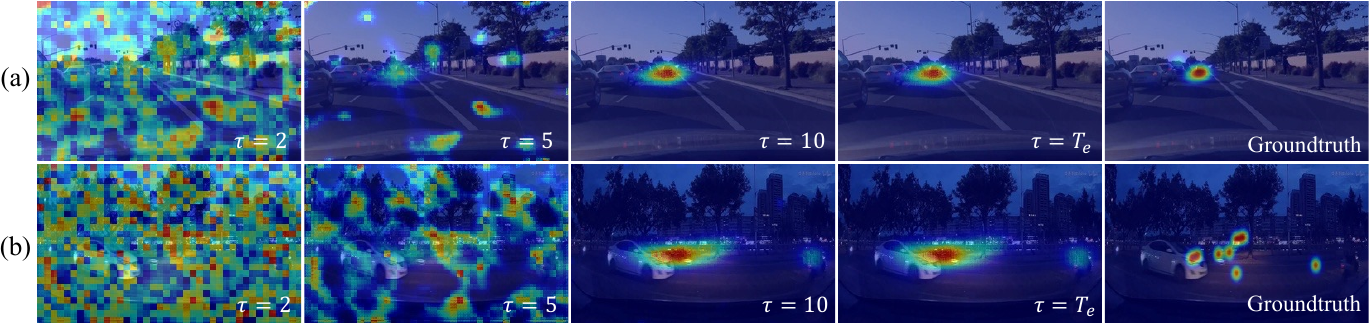}
    \caption{Visualization of the denoising process ($\tau$: current step).}
    \label{fig::denoise_step_sample}
\end{figure}

\textbf{Ablation study on denoising process.} To further investigate the refinement process of denoising, we visualize intermediate results in Fig. \ref{fig::denoise_step_sample}. As illustrated, our model converges rapidly within only a few steps, transitioning from random noise to dispersed but plausible clusters of attention, and ultimately forming a human-like attention distribution. 
\par
\begin{figure}[t]
    \centering
    \includegraphics[width=1\columnwidth]{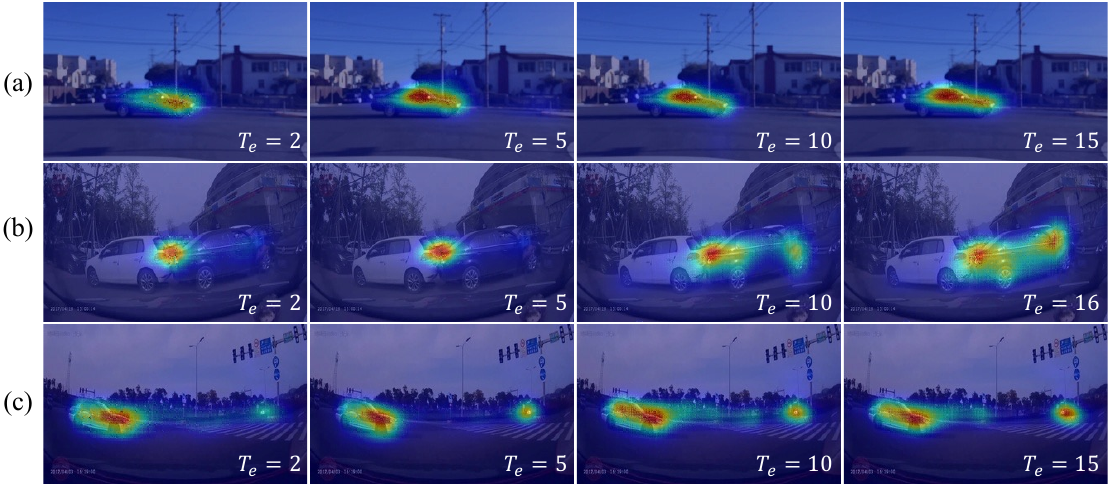}
    \caption{Visualization of denoising results with different $T_e$.}
    \label{fig::denoise_sample}
\end{figure}

 Meanwhile, to examine the effect of denoising steps more systematically, we conducted ablation experiments on the DADA-2000 and BDD-A datasets. Specifically, we fix the total diffusion steps $T_i$ to 300 for both datasets (same as default), and consider two strategies: (1) training the model with different denoising step settings, and (2) altering the number of denoising steps only during inference without retraining. The quantitative results reported in Table \ref{tab::ablation_denoise} indicate that directly reducing the number of denoising steps at inference leads to severe performance degradation. In contrast, when the model is trained with reduced denoising steps, it maintains relatively strong performance. Although such configurations do not surpass the default experimental setting, they still achieve competitive results compared with some baselines, even under a small number of denoising steps. Representative qualitative results are shown in Fig. \ref{fig::denoise_sample}, which further highlight that our model can already reach quantitatively competitive performance with as few as 2$\sim$5 denoising steps. Nevertheless, for the main experiments we adopt 15 or 16 denoising steps, as this setting yields consistently superior results for SoTA benchmarks. Further exploration is required regarding model deployment, since increasing the number of denoising steps inevitably leads to higher GPU consumption and reduced inference speed. Nevertheless, under the current configuration, an inference speed of 8 FPS remains practically feasible.
\input{tables/ablation_denoising}

%% file: tables/result_all.tex
\begin{table*}[t]
\caption{\small Test performance on four datasets (``I'' indicates single image input. ``V'' implies video sequence. ``S'' indicates single modal of RGB image. ``M'' indicates multiple modals of input such as semantic segmentation. ``\#'' indicates number of computation parameters. Results best in \textbf{bold}, second best \underline{underlined}. ``-'' implies metrics not reported. ``N/A'' indicates no implementation due to data requirement.) }
\centering
\setlength{\tabcolsep}{0.3pt}
\begin{tabular}{ccc|*{5}{c}|*{5}{c}|*{3}{c}|*{5}{c}}
\toprule
\multicolumn{3}{c}{Dataset} \vline & \multicolumn{5}{c}{\cellcolor{cyan!8}\textbf{TrafficGaze}}\vline & \multicolumn{5}{c}{\cellcolor{green!8}\textbf{DADA-2000}}\vline & \multicolumn{3}{c}{\cellcolor{blue!8}\textbf{BDD-A}}\vline & \multicolumn{5}{c}{\cellcolor{purple!8}\textbf{DrFixD-rainy}}\\
\midrule
\multicolumn{1}{c}{Method}\vline & \multicolumn{1}{c}{Input} & \multicolumn{1}{c}{\#.$\downarrow$}\vline & $\textit{KLD}\downarrow$ & $\textit{CC}\uparrow$ & $\textit{SIM}\uparrow$ & $\textit{NSS}\uparrow$ & $\textit{AUC-J}\uparrow$ & $\textit{KLD}\downarrow$ & $\textit{CC}\uparrow$ & $\textit{SIM}\uparrow$ & $\textit{NSS}\uparrow$ & $\textit{AUC-J}\uparrow$ & $\textit{KLD}\downarrow$ & $\textit{CC}\uparrow$ & $\textit{SIM}\uparrow$ & $\textit{KLD}\downarrow$ & $\textit{CC}\uparrow$ & $\textit{SIM}\uparrow$ & $\textit{NSS}\uparrow$ & $\textit{AUC-J}\uparrow$\\
\hline
\multicolumn{1}{c}{MLNet \cite{MLNet}}\vline & I+S & 15M & 0.87 & 0.87 & 0.45 & 5.69 & 0.90 & 11.78 & 0.04 & 0.07 & 0.30 & 0.59 & 1.20 & 0.61 & 0.43 & 3.69 & 0.79 & 0.63 & 3.90 & 0.93\\

\multicolumn{1}{c}{ACLNet \cite{ACLNet}}\vline & V+S & - & 0.27 & 0.91 & 0.77 & 5.73 & 0.95 & 1.95 & 0.46 & 0.30 & 3.23 & 0.93 & 1.12 & 0.61 & 0.46 & - & - & - & - & -\\
\multicolumn{1}{c}{CPFE \cite{CPFE}}\vline & I+S & - & 0.26 & 0.89 & 0.77 & 4.33 & 0.94 & 2.18 & 0.33 & 0.21 & 2.43 & 0.91 & 1.65 & 0.42 & 0.30 & - & - & - & - & - \\

\multicolumn{1}{c}{TASED-Net \cite{TASED-Net}}\vline & V+S & 21M & 1.43 & 0.94 & 0.79 & 5.73 & 0.97 & 1.78 & 0.46 & 0.31 & 3.20 & 0.92 & 1.24 & 0.55 & 0.42 & 0.85 & 0.84 & 0.59 & 4.21 & 0.95\\
\multicolumn{1}{c}{CDNN \cite{CDNN}}\vline & I+S & $<$\underline{1M} & 0.29 & \underline{0.95} & 0.78 & 5.83 & 0.97 & 1.83 & 0.43 & 0.31 & 2.93 & \underline{0.94} & 1.14 & 0.62 & 0.45 & 0.52 & 0.82 & 0.63 & 4.11 & 0.95\\
\multicolumn{1}{c}{SCAFNet \cite{DADA}}\vline & V+M & 55M & 0.66 & 0.94 & 0.77 & 6.10 & \underline{0.98} & 2.19 & 0.50 & \textbf{0.37} & 3.34 & 0.92 & 1.48 & 0.56 & 0.40 & 1.87 & 0.84 & 0.67 & 4.17 & 0.94\\
\multicolumn{1}{c}{DrFixD-rainy \cite{drfixd}}\vline & V+S & 40M & 0.28 & 0.94 & 0.78 & 6.01 & \underline{0.98} & 1.78 & 0.45 & 0.29 & 3.00 & \underline{0.94} & 1.09 & \textbf{0.64} & 0.47 & 0.47 & 0.85 & 0.67 & 4.19 & \underline{0.96}\\
\multicolumn{1}{c}{FBLNet \cite{fblnet}}\vline &  I+S & 87M & 0.46 & 0.90 & 0.69 & \textbf{6.50} & 0.97 & 1.92 & 0.50 & 0.33 & \textbf{4.13} & \textbf{0.95} & 1.40 & \textbf{0.64} & 0.47 & 0.50 & 0.85 & 0.69 & 4.29 & 0.95 \\
\multicolumn{1}{c}{SCOUT+ \cite{scout}}\vline & V+M & 5M & 0.39 & 0.91 & 0.72 & 5.35 & 0.97 & N/A & N/A & N/A & N/A & N/A & \textbf{1.04} & \underline{0.63} & 0.48 & 0.50 & 0.84 & 0.65 & 3.95 & 0.95\\
\multicolumn{1}{c}{SalM$^2$ \cite{salm2}}\vline & I+S & $<$\textbf{0.1M} & 0.28 & 0.94 & 0.78 & 5.90 & \underline{0.98} & 1.71 & 0.37 & 0.31 & 2.10 & 0.92 & \underline{1.08} & \textbf{0.64} & 0.47 & 0.47 & \underline{0.86} & 0.68 & 4.31 & 0.95\\
\multicolumn{1}{c}{MTSF \cite{mtsf}}\vline & V+S & 194M & 0.26 & \textbf{0.96} & \textbf{0.82} & 5.98 & \underline{0.98} & \underline{1.61} & \underline{0.51} & \underline{0.36} & 3.44 & 0.93 & 1.61 & 0.51 & 0.36 & 0.62 & 0.77 & 0.61 & 4.02 & \underline{0.96}\\
\multicolumn{1}{c}{DPSNN \cite{DPSNN}}\vline & I+S & 8M & \underline{0.25} & \underline{0.95} & \underline{0.80} & 6.07 & \underline{0.98} & 1.84 & 0.43 & 0.30 & 2.89 & \underline{0.94} & 1.52 & 0.53 & 0.43 & \underline{0.41} & 0.85 & \underline{0.71} & \underline{4.39} & \textbf{0.97}\\
\multicolumn{1}{c}{SalEMA \cite{SalEMA}}\vline & V+S & 218M & 0.30 & 0.93 & 0.77 & 5.81 & 0.97 & 1.65 & 0.48 & 0.33 & 3.26 & \textbf{0.95} & 1.20 & 0.60 & 0.45 & 0.47 & 0.85 & 0.67 & 4.11 & 0.95\\
\multicolumn{1}{c}{Gate-DAP \cite{gate}}\vline & V+M & 106M & 0.33 & 0.92 & 0.77 & 5.92 & \underline{0.98} & 1.65 & \textbf{0.52} & \underline{0.36} & 3.14 & - & 1.12 & 0.61 & \underline{0.49} & 0.48 & 0.85 & 0.69 & 4.21 & \textbf{0.97}\\

\multicolumn{1}{c}{Fu \textit{et al.} \cite{fu}}\vline & I+S & 41M & 0.26 & 0.93 & 0.73 & - & - & 2.30 & 0.47 & 0.34 & 3.21 & 0.91 & - & - & - & - & - & - & - & -\\

\midrule
\multicolumn{1}{c}{\textbf{DiffAttn (Ours)}}\vline & I+S & 92M & \textbf{0.24} & \underline{0.95} & \textbf{0.82} & \underline{6.11} & \textbf{0.99} & \textbf{1.58} & \underline{0.51} & \underline{0.36} & \underline{3.49} & \textbf{0.95} & 1.09 & \textbf{0.64} & \textbf{0.50} & \textbf{0.40} & \textbf{0.88} & \textbf{0.73} & \textbf{4.59} & \textbf{0.97}\\
\bottomrule
\label{tab::result_all}
\end{tabular}
\end{table*}

%% file: tables/ablation_LLM.tex
\begin{table}[t]
\caption{\small Ablation study on LLM-based semantic enhancement (``F'' and ``T'' denote frozen and trainable parameters, respectively).}
\centering
\setlength{\tabcolsep}{0.9pt}
\begin{tabular}{*{2}{c}|*{3}{c}|*{5}{c}}
\toprule
\multicolumn{2}{c}{Dataset}\vline & \multicolumn{3}{c}{\textbf{BDD-A}}\vline & \multicolumn{5}{c}{\textbf{DADA-2000}}\\
\midrule
\multicolumn{1}{c}{Method}\vline & \multicolumn{1}{c}{\#.$\downarrow$}\vline & $\textit{KLD}\downarrow$ & $\textit{CC}\uparrow$ & $\textit{SIM}\uparrow$ & $\textit{KLD}\downarrow$ & $\textit{CC}\uparrow$ & $\textit{SIM}\uparrow$ & $\textit{NSS}\uparrow$ & $\textit{AUC-J}\uparrow$\\
\midrule
\multicolumn{1}{c}{\cellcolor{gray!15}w/o $\mathcal{F}_\text{LLM}$} \vline & \textbf{91M} &1.126 & 0.632 & \textbf{0.506} & 1.639 & 0.502 & 0.377 & 3.472 & \textbf{0.949}\\
\multicolumn{1}{c}{\cellcolor{cyan!8}\textbf{LLaMA-F}}\vline & \underline{92M} & \textbf{1.087} & \textbf{0.636} & \underline{0.502} & \textbf{1.582} & \textbf{0.510} & 0.360 & \underline{3.486} & \textbf{0.949}\\
\multicolumn{1}{c}{\cellcolor{red!8}LLaMA-T}\vline & 153M & 1.114 & 0.632 & 0.501 & \underline{1.606} & \underline{0.507} & 0.370 & \textbf{3.493} & \textbf{0.949} \\
\multicolumn{1}{c}{\cellcolor{cyan!8}DeepSeek-F}\vline & \underline{92M} & \underline{1.097} & \underline{0.634} & 0.497 & 1.660 & 0.502 & \textbf{0.381} & 3.474 & \textbf{0.949} \\
\multicolumn{1}{c}{\cellcolor{red!8}DeepSeek-T}\vline & 139M & 1.113 & 0.629 & 0.494 & 1.684 & 0.495 & \underline{0.378} & 3.427 & \underline{0.948}\\
\bottomrule
\end{tabular}
\label{tab::ablation_LLM}
\end{table}

%% file: tables/ablation_denoising.tex
\begin{table}[t]
\caption{\small Ablation study on denoising steps $T_e$.}
\renewcommand{\arraystretch}{0.9}
\centering
\setlength{\tabcolsep}{2.6pt}
\begin{tabular}{c|*{5}{c}|*{3}{c}}
\toprule
\multicolumn{1}{c}{Dataset}\vline & \multicolumn{5}{c}{\textbf{DADA-2000} $(T_e=16)$}\vline & \multicolumn{3}{c}{\textbf{BDD-A} $(T_e=15)$} \\
\midrule
\multicolumn{1}{c}{Method}\vline & $\textit{KLD}\downarrow$ & $\textit{CC}\uparrow$ & $\textit{SIM}\uparrow$ & $\textit{NSS}\uparrow$ & $\textit{AUC-J}\uparrow$ & $\textit{KLD}\downarrow$ & $\textit{CC}\uparrow$ & $\textit{SIM}\uparrow$ \\
\midrule
\multicolumn{1}{c}{$\boldsymbol{\tau=T_e}$}\vline & \textbf{1.582} & \textbf{0.510} & 0.360 & \underline{3.486} & \underline{0.949} & \textbf{1.087} & \textbf{0.636} & \textbf{0.502} \\
\midrule
\multicolumn{9}{c}{\cellcolor{gray!13}(1) Train model with different denoising steps} \\
\midrule
\multicolumn{1}{c}{$T_e=10$}\vline & 1.595 & 0.507 & 0.361 & 3.465 & \underline{0.949} & 1.100 & 0.634 & \textbf{0.502} \\
\multicolumn{1}{c}{$T_e=5$}\vline & 1.606 & 0.506 & \textbf{0.374} & \textbf{3.488} & \textbf{0.950} & \underline{1.090} & \underline{0.635} & \underline{0.500} \\
\multicolumn{1}{c}{$T_e=2$}\vline & \underline{1.586} & \underline{0.509} & \underline{0.366} & \underline{3.486} & \textbf{0.950} & 1.096 & 0.634 & \underline{0.500} \\
\midrule
\multicolumn{9}{c}{\cellcolor{gray!13}(2) Change denoising steps without training model} \\
\midrule
\multicolumn{1}{c}{$\tau=10$}\vline & 1.626 & 0.505 & 0.368 & 3.451 & 0.948 & 3.065 & 0.099 & 0.108\\
\multicolumn{1}{c}{$\tau=5$}\vline & 1.926 & 0.447 & 0.268 & 3.034 & 0.930 & 2.687 & 0.161 & 0.133\\
\multicolumn{1}{c}{$\tau=2$}\vline &  3.472 & 0.016 & 0.060 & 0.117 & 0.612 & 3.096 & 0.037 & 0.093\\
\bottomrule
\end{tabular}
\label{tab::ablation_denoise}
\end{table}